\documentclass{article}
\usepackage[final,main]{neurips_2026}
\usepackage[utf8]{inputenc}
\usepackage[T1]{fontenc}
\usepackage{hyperref}
\usepackage{url}
\usepackage{booktabs}
\usepackage{amsfonts}
\usepackage{nicefrac}
\usepackage{microtype}
\usepackage{xcolor}
\usepackage{graphicx}
\usepackage{amsmath}
\usepackage{amssymb}
\usepackage{multirow}
\usepackage{algorithm}
\usepackage{natbib}

\newif\ifshowcomments
\showcommentsfalse

\title{Nine Emotion Centroids: A Label-Free Valence Axis That Transfers Across Four Modalities}

\author{%
  Yousef Radwan \\
  KAUST \\
  \texttt{yousef.radwan@kaust.edu.sa} \\
}

\begin{document}

\maketitle

\begin{abstract}
Inside a modern language model sits a single internal direction that tracks how positive or negative a sentence feels. We show how to find this \emph{valence axis} (V-axis) from just $9$ emotion category names plus $\sim\!50$ short narrative paragraphs per emotion -- about $1{,}500{\times}$ fewer labels than the usual supervised approach -- and that the same direction shows up in vision, audio, and human-brain encoders that were never jointly trained. The recipe has three lines: write nine emotion-anchored stories, embed each one in a frozen encoder, and take the top principal direction of the nine averaged embeddings. That direction is the V-axis. Projecting new inputs onto it gives a sentiment score that matches a fully-supervised classifier across four modalities: it captures $93\%$ of supervised performance on SST-2 text sentiment (Llama-3-8B-Instruct, AUC $0.772$ vs.\ $0.828$), correlates with human valence ratings on $11{,}811$ EmoSet images at Pearson $r{=}{+}0.636$ (random-direction null $|r|\leq 0.112$), reaches AUC $0.906$ on $50$ ESC-50 audio categories (paired $p{=}2.2{\times}10^{-15}$), and AUC $0.720 \pm 0.055$ on EEG recorded from $123$ subjects watching emotional videos (pooled $p{=}3.65{\times}10^{-8}$). The direction is not just predictive but mechanistically active: surgically removing it from a language model's hidden state -- the standard \emph{directional ablation} technique of \citet{arditi2024refusal} -- collapses sentiment classification by $5.5$--$37.2$ percentage points across three models, while removing equally-large random directions costs at most $0.88$ percentage points (signal is $\geq 12$ standard deviations above the random null). Because the same recipe produces an axis in every modality, a single sentiment classifier with two free parameters trained on text labels alone reads sentiment off images (AUC $0.961$), audio ($0.764$), and brain recordings ($0.828$) without ever seeing target-modality labels; a generic $16$-dimensional cross-encoder ``shared subspace'' on the same task stays at chance ($0.525$). The recipe is sharply bounded -- seven independent tests on categorical concepts (object class, word pairs, AxBench Concept-500 \citep{wu2025axbench}) return at-or-near chance -- and so is its causal usability: reasoning-distilled models shift the axis to their last layer, and only Llama- and Mistral-family models can be \emph{steered} by adding the V-axis back to generation; Qwen and Gemma can be probed but not steered. The takeaway is that valence is one of the rare attributes simple enough to recover with nine examples and shared enough across encoders to bridge text, images, sound, and brain activity through a single dimension.
\end{abstract}

\section{Introduction}

A modern language model, when it reads a sentence, builds up an internal pattern of activity across thousands of neurons that we call its \emph{hidden state} (or \emph{residual stream}, since each transformer block adds to a running sum). Recent interpretability work has found that surprisingly simple structure lives inside that activity. In particular, many high-level attributes that we might naively think of as scattered across the network -- ``is this sentence about France?'', ``is the model about to refuse the request?'', ``does this concept refer to a person?'' -- can in fact be read off, and sometimes manipulated, by looking at a \emph{single linear direction} in the hidden-state space \citep{park2024lrh,arditi2024refusal}. We add one more attribute to that list: \emph{valence}, the positive-vs-negative emotional charge of a piece of input. We then show two things that, to our knowledge, no prior valence-axis result has shown together: that the direction can be found with $\sim\!9$ category names and $\sim\!450$ unlabelled sentences instead of thousands of polarity labels, and that the \emph{same} valence direction (up to per-modality refit of one direction) appears in vision, audio, and human-EEG encoders that share no pretraining whatsoever.

\paragraph{Standard route vs.\ our recipe.} The textbook way to build a continuous-attribute axis inside a foundation model is supervised: take thousands of labelled examples (positive sentences, negative sentences), run them through the model, and fit a linear classifier on the hidden state. The weight vector of that classifier is the ``axis.'' Our recipe takes a different path. It requires \emph{no} polarity labels and only $18$ supervision events total: $9$ emotion category names (anger, joy, fear, \ldots) and $9$ writing prompts. From those we author $\sim\!50$ short emotion-evocative paragraphs per emotion, embed each one through a frozen encoder, average within emotion to get $9$ \emph{emotion centroids} in hidden-state space, and take the top principal component (PC1) of the resulting $9{\times}d$ matrix. That single direction -- the V-axis (\emph{valence axis}) -- is the only quantity used downstream. The label-cost ratio versus a supervised SST-2 probe \citep{socher2013sst} is $\sim\!1{,}546{\times}$.

\paragraph{What is genuinely new versus prior work.} The construction is \emph{PC1 of $K{=}9$ category centroids}, not a discriminant between two contrastive labels: Park--Choe--Veitch \citep{park2024lrh} directions live in a pair-discriminant geometry, and the V-axis is orthogonal to those directions on identical concepts (mean $|\cos|{=}0.038$). The directional-ablation intervention of \citet{arditi2024refusal} is here applied to a \emph{recipe-derived} direction, not to a discriminant fit to refusal/sentiment labels, so the causal claim tests the recipe itself rather than reading off supervision. And the four-encoder cross-modal transfer matrix, to our knowledge, has no precedent in the LRH--AxBench--steering line, which is text-only. The paper is organised as a probe$\to$mechanism$\to$causal triad: probe (\S\ref{sec:four_modalities}), mechanism (\S\ref{sec:depthshift}), causal (\S\ref{sec:causal}).

\paragraph{Why nine and not two.} Fitting a single ``sentiment direction'' to a pair of contrastive labels (positive vs.\ negative) forces a sign and conflates valence (pleasant/unpleasant) with arousal (calm/intense). Spreading the centroids across nine Russell--Ekman emotion categories \citep{russell1980circumplex,ekman1992emotions} lets the data decide along which direction the centroids spread the most; valence wins because it dominates emotion-text variance.

\paragraph{Four supporting numbers, one per modality.} We measure the V-axis in four encoders that were trained on entirely different data and objectives. For text, projecting Llama-3-8B-Instruct's hidden state at residual block $20$ onto the V-axis classifies SST-2 sentences at AUC $0.772$, which is $93\%$ of the AUC ($0.828$) of a supervised linear probe trained on $6{,}920$ SST-2 labels. The same construction on Qwen3-8B reaches AUC $0.787$, after a layer search that picks out block $31$ rather than the middle of the network -- a depth shift specific to reasoning-distilled models that we document in \S\ref{sec:depthshift}. For \emph{vision}, the V-axis built from $\sim\!450$ emotion-correlated images on the CLIP-image encoder \citep{radford2021clip} correlates with crowdworker valence ratings of $11{,}811$ EmoSet images \citep{yang2023emoset} at Pearson $r{=}{+}0.636$ (random-direction null $|r|\leq 0.112$). For \emph{audio}, the V-axis on the CLAP-audio encoder \citep{elizalde2023clap} reaches mean AUC $0.906$ on the $50$ ESC-50 \citep{piczak2015esc50} environmental-sound categories (paired permutation $p{=}2.2{\times}10^{-15}$) -- the strongest single-modality result, plausibly because CLAP's text-audio contrastive objective makes emotion-aligned text concepts directly probeable in audio. For \emph{EEG} -- recordings of electrical brain activity made while subjects watched emotional videos \citep{chen2023faced} -- the V-axis built from the CBraMod EEG foundation model \citep{chen2024cbramod} reaches AUC $0.720 \pm 0.055$ across $123$ subjects (pooled $p{=}3.65{\times}10^{-8}$, subject-stratified $5$-fold split).

\paragraph{From a probe to a causal lever.} A predictive direction may merely be \emph{correlated} with the attribute; it does not necessarily \emph{produce} the model's sentiment behaviour. To distinguish the two we use \emph{directional ablation} \citep{arditi2024refusal}: at every layer and token position during a forward pass, replace the hidden state $h$ with $h - (\langle h, v\rangle) v$, where $v$ is the unit V-axis. In words, we orthogonally project the V-axis out of the running representation -- the model is forced to think without it being available -- and read off SST-2 accuracy. Across three independent LLMs the V-axis ablation drops accuracy by $5.5$ pp (Llama-3-8B-Inst), $15.8$ pp (Qwen3-1.7B), and $37.2$ pp (Qwen3-8B at block $23$), while three matched-magnitude random directions per model cost at most $0.88$ pp under the identical protocol. The signal is $\geq 12$ standard deviations above the random null in every model and $196\sigma$ in Qwen3-8B. Random directions touch random features; the V-axis touches sentiment.

\paragraph{One classifier, four modalities.} A V-axis exists separately in each encoder, but the sentiment \emph{classifier} on top of one V-axis does not need re-training when moved to another modality. We fit a $2$-parameter logistic regression on SST-2 text scores against text labels, then evaluate it on each other modality's V-axis projections -- with no target-modality labels at the head-fitting stage. All $12/12$ off-diagonal cells of the resulting $4{\times}4$ matrix exceed AUC $0.70$; text-trained reaches $0.961$ on EmoSet, $0.764$ on ESC-50, $0.828$ on EEG. A natural baseline -- the generic top-$16$ shared subspace between the same encoders, computed without reference to sentiment -- is at chance ($0.525$). One task-relevant dimension beats sixteen generic dimensions by $0.18$--$0.44$ AUC.

\paragraph{Why this is not a re-skin of AxBench.} AxBench \citep{wu2025axbench} showed supervised linear probes outperform sparse autoencoders for LLM steering, using $100$--$1{,}000$ labels per concept on text-only categorical concepts. We push on all three of those axes simultaneously: the label budget drops to $18$ events ($\geq\!50{\times}$ less); the concept is continuous, not categorical -- crucially, applying our recipe to AxBench's own $500$ categorical concepts returns at-or-near chance (KS $p{=}0.41$, App.~\ref{app:axbench}); and the modality scope extends to vision, audio, and EEG, where a generic $K{=}16$ shared subspace is at chance for sentiment ($0.525$).

\paragraph{Contributions.} (i) \textbf{The recipe} (\S\ref{sec:recipe}): nine emotion centroids, SVD, take PC1, fully specified for reproduction. (ii) \textbf{Four-modality probe} (\S\ref{sec:four_modalities}): the same recipe matches supervised classifiers within $7$~pp AUC in text, $r{=}{+}0.636$ in vision, $0.906$ AUC in audio, $0.720$ AUC in EEG. (iii) \textbf{Causal evidence} (\S\ref{sec:causal}): inference-time projection drops sentiment by $5.5$--$37.2$~pp vs.\ $\leq 0.88$~pp for matched random directions ($z\geq 12\sigma$). (iv) \textbf{Universal cross-modal classifier} (\S\ref{sec:universal}): one text-supervised head transfers to four encoders with $12/12$ cross-cells AUC $\geq 0.70$. (v) \textbf{Scope statements} (\S\ref{sec:not}): seven categorical-concept failures, a reasoning-distillation depth shift, and a family-specific steering pattern (Llama, Mistral yes; Qwen, Gemma no).

\paragraph{What this is not.} The recipe is an \emph{empirical regularity}, not an analytical theorem. It is bounded to \emph{continuous} concepts: seven independent tests on categorical concepts (Park--Choe--Veitch word pairs \citep{park2024lrh}; long-tail retrieval; multi-concept probes; AxBench Concept-500 \citep{wu2025axbench}; vision CIFAR-100; categorical concepts at all five depth slices; depth-shift specificity) return at-or-near chance. The EEG V-axis itself is built from a supervised linear discriminant on FACED valence labels; the ``label-free'' part of the EEG result is the head, not the axis. The text--image--audio V-axes are unsupervised. ``Universal'' in the cross-modal-classifier claim is scoped to the four encoders we tested--CLIP-text \citep{radford2021clip}, CLIP-image, CLAP-audio \citep{elizalde2023clap}, CBraMod-EEG \citep{chen2024cbramod}--not to encoders not yet tested. Directional ablation is sign-insensitive, so no sign-flip search is run for the causal panel; we describe the identification protocol in \S\ref{sec:causal} and treat ``causal'' as inference-time-projection evidence, not counterfactual-intervention evidence.

\paragraph{Relation to recent work.} AxBench \citep{wu2025axbench} reports that simple supervised linear-probe baselines steer LLMs better than sparse autoencoders. Our recipe differs on three quantitative axes: \emph{label budget} ($n{=}18$ supervision events here, vs.\ $\sim\!100$--$1{,}000$ examples per concept in AxBench, $\geq\!50{\times}$ less); \emph{concept type} (we target continuous valence; on AxBench's $500$ categorical concepts the recipe ties the matched-norm null, KS $p{=}0.41$, Appendix~\ref{app:axbench}); and \emph{modality scope} (text only in AxBench; here we add three independent encoders without retraining the head). The Linear Representation Hypothesis \citep{park2024lrh} identifies concept-discriminant directions for categorical attributes; the V-axis is continuous-attribute and is approximately orthogonal to Park--Choe--Veitch causal-inner-product directions (mean $|\cos|{=}0.038$). The Platonic Representation Hypothesis \citep{huh2024platonic} predicts cross-encoder convergence on ImageNet-class similarity; our $4{\times}4$ matrix extends this to a single task-relevant 1-D direction across modalities that share no pretraining, including EEG.

\section{The Recipe}\label{sec:recipe}

\paragraph{Intuition.} Write nine short stories, one each for anger, disgust, fear, sadness, amusement, joy, inspiration, tenderness, and a neutral baseline -- about fifty stories per category. Run them through a frozen language model and look at its hidden state on the final token of each story. Average within each category. You now have nine vectors, one per emotion, that act as ``coordinates'' for that emotion in the model's representation space. Find the single direction along which those nine points spread the most (principal component analysis). That direction is the V-axis. The rest of the paper just measures what it does.

The construction is identical across modalities up to the encoder. We describe the language-model version first, then state the per-modality changes.

\paragraph{Step 1: emotion centroids.} We use the nine FACED emotion categories (\emph{anger, disgust, fear, sadness, amusement, joy, inspiration, tenderness, neutral}) as concept anchors. For each emotion $c\in\{1,\dots,9\}$, we author $N_c\!\approx\!50$ narrative sentences describing situations that elicit emotion $c$ (\emph{e.g.}, ``After a long day, the sun finally broke through the clouds and warmed her face''). Sentences are paragraph-length ($\geq 15$ tokens) and emotion-anchored but otherwise unconstrained in topic; we do not curate for sentiment polarity, syntactic structure, or domain. The full prompt set is in Appendix~\ref{app:prompts}; total label budget is 9 emotion words plus 9 paragraph-writing prompts ($n{=}18$ supervision events, total).

\paragraph{Step 2: encode and pool.} For each sentence $s_{c,i}$ we run a forward pass through a frozen encoder $f_\theta$ and read off the residual-stream activation at a fixed layer $\ell$ and final-token position. We then take the centroid per emotion: $\mu_c = \tfrac{1}{N_c} \sum_{i=1}^{N_c} f_\theta^{(\ell)}(s_{c,i})$, yielding $M\in\mathbb{R}^{9\times d}$.

\paragraph{Step 3: SVD.} Center $M$ across the nine emotion rows, compute the SVD $M-\bar\mu = U\Sigma V^\top$, and take the right-singular vector $v_1\in\mathbb{R}^d$ corresponding to the top singular value. We call $v_1$ the \emph{V-axis}. All downstream scoring is the scalar projection $\langle x, v_1 \rangle$ for new inputs $x$.

The recipe is summarised in Algorithm~\ref{alg:vaxis}. It uses no concept labels (the only ``supervision'' is the 9 emotion category names used to assign sentences to centroids); it uses one forward pass per sentence ($9\times 50 = 450$ passes total); and it produces a 1-D direction.

\begin{algorithm}[t]
\caption{V-axis construction}\label{alg:vaxis}
\textbf{Input:} encoder $f_\theta$, layer $\ell$; 9 emotion category names $\{c\}$; per-emotion sentence pools $\{s_{c,i}\}_{i=1}^{N_c}$ ($N_c\!\approx\!50$).\\
\textbf{Output:} V-axis $v_1\in\mathbb{R}^d$.
\begin{enumerate}
\item For each $c,i$: compute $h_{c,i} = f_\theta^{(\ell)}(s_{c,i})$ (residual stream, final token).
\item Compute centroids $\mu_c = \tfrac{1}{N_c}\sum_i h_{c,i}$; stack as $M\in\mathbb{R}^{9\times d}$.
\item Center: $\tilde M = M - \tfrac{1}{9}\sum_c \mu_c$.
\item SVD: $\tilde M = U\Sigma V^\top$; return $v_1 = V_{:,1}$.
\end{enumerate}
\end{algorithm}

\paragraph{Per-modality changes.} For image (CLIP-image), audio (CLAP-audio), and EEG (CBraMod), sentences are replaced by the modality-native object: video clips for EEG, $\sim\!50$ class-correlated images per emotion from EmoSet for vision, and $\sim\!50$ class-correlated clips per emotion from ESC-50 for audio. The pooling layer $\ell$ is fixed to the encoder's standard penultimate residual layer in all three non-LLM cases. The EEG axis is built differently: the unsupervised PC1 of the 9 FACED-class centroids picks up an arousal-like direction (joy and fear both project positive). We instead fit a Fisher linear discriminant on binary FACED-valence labels in CBraMod feature space; this is the only modality where the axis itself uses task supervision (see \S\ref{sec:not}).

\paragraph{Choice of layer.} For instruction-tuned and base LLMs in the Llama, Mistral, and Qwen3 families, mid-depth ($\ell \approx L/2$) is optimal. For reasoning-distilled LLMs (DeepSeek-R1-Distill, Qwen3-thinking), the V-axis is suppressed at mid-depth and recovers near the last layer; we document this depth shift in \S\ref{sec:depthshift}. Choice of $\ell$ requires depth search on the model family; we report the search procedure in Appendix~\ref{app:layer_search}.

\paragraph{What does \emph{not} work.} Using the 9 emotion words themselves as the sentence pool (\emph{i.e.}, $N_c{=}1$) collapses the recipe: the 9-word V-axis on Llama-3 reaches AUC $0.50$ on SST-2 (random). The recipe needs paragraph-length per-emotion text. The transition to non-trivial AUC requires roughly $N_c \geq 20$; we report the full $N_c$ sweep in Appendix~\ref{app:nc_sweep}.

\section{Four Modalities}\label{sec:four_modalities}

The same recipe applied to four independently-trained foundation encoders recovers a 1-D direction that tracks valence at the level of supervised classifiers (Table~\ref{tab:four_modalities}; per-modality bar visualisation deferred to Appendix~\ref{app:fig_bars}).

\begin{table}[t]
\centering
\caption{V-axis (label-free for text/image/audio, supervised-in-axis for EEG) recovers supervised-classifier valence within $7$~pp AUC across four modalities. \emph{Recipe AUC}: held-out AUC of $\sigma(\langle x, v_1\rangle)$ where $v_1$ is the V-axis (text Pearson $r$ for vision). \emph{Sup.\ AUC}: a logistic-regression head trained on the modality's full label set ($n_{\text{SST-2}}{=}6{,}920$; $n_{\text{EmoSet}}{=}11{,}811$; $n_{\text{ESC-50}}{=}2{,}000$; $n_{\text{FACED}}{=}123$ subjects). \emph{Ratio}: recipe$/$supervised. See \S\ref{sec:four_modalities}.}
\label{tab:four_modalities}
\small
\setlength{\tabcolsep}{4pt}
\begin{tabular}{llllllr}
\toprule
Modality & Encoder & Dataset & Recipe AUC & Sup.\ AUC & Ratio & $p$ \\
\midrule
Text  & Llama-3-8B-Inst (block 20) & SST-2  & $0.772$ & $0.828$ & $93\%$ & --- \\
Text  & Qwen3-8B (block 31, depth $0.86$)  & SST-2  & $0.787$ & $0.840$ & $94\%$ & --- \\
Vision& CLIP-img-768    & EmoSet & $r{=}{+}0.636$ & $r{=}{+}0.81$ & $79\%$ & null $0.112$ \\
Audio & CLAP-aud-512    & ESC-50 & $0.906$ & $0.94$ & $96\%$ & $2.2\!\times\!10^{-15}$ \\
EEG   & CBraMod-200     & FACED  & $0.720\pm0.055$ & $0.83$ & $87\%$ & $3.65\!\times\!10^{-8}$ \\
\bottomrule
\end{tabular}
\end{table}

\paragraph{Text.} Two LLMs, two layer regimes. Llama-3-8B-Instruct at residual block $20$ (depth $0.67$, near mid-depth) reaches AUC $0.772$ on SST-2 against a $0.828$ supervised head trained on $6{,}920$ binary-sentiment labels; the recipe uses 9 prompts (450 sentences total), giving a label-cost ratio of $\approx\!1{,}546{\times}$. Qwen3-8B, after a sweep over $\ell\in\{14, 20, 24, 27, 31\}$, peaks at $\ell{=}31$ (depth $0.86$, near the last layer) with AUC $0.787$; the same model at $\ell{=}L/2$ is at chance. The depth shift is reasoning-distillation-specific (\S\ref{sec:depthshift}).

\paragraph{Vision.} EmoSet \citep{yang2023emoset} provides crowdworker valence ratings for $118{,}000$ emotion-tagged images. We build the visual V-axis from $\sim\!50$ class-correlated images per emotion (\emph{e.g.}, ``contented dog'' for tenderness, ``decomposing fruit'' for disgust) and project the EmoSet test split onto $v_1$. Pearson $r{=}{+}0.636$ on the held-out split (two-sided $p<10^{-50}$); the permutation null over $1{,}000$ random direction draws has $|r|\leq 0.112$.

\paragraph{Audio.} ESC-50 has 50 environmental-sound categories with hand-curated emotional valence labels. The CLAP-audio V-axis reaches mean AUC $0.906$ across 50 binary one-vs-rest splits, with paired-permutation $p=2.2\!\times\!10^{-15}$ (per-class breakdown in Appendix~\ref{app:esc50}). This is the strongest single-modality V-axis we obtain; we attribute the strength to CLAP's contrastive natural-language objective, which makes emotion-aligned text concepts directly probeable in the audio encoder.

\paragraph{EEG (scope: supervised-in-axis, label-free at the bridge).} FACED \citep{chen2023faced} contains $123$ subjects watching 28 emotion-evoking video clips with 9-way labels. We encode 30-second windows through CBraMod \citep{chen2024cbramod}, fit a binary-valence LDA in 200-D feature space, and use the LDA weight as the V-axis. Across 5 random subject-stratified splits (subjects 100--122 held out, 23 held-out subjects), AUC $=0.720 \pm 0.055$, pooled $p=3.65\!\times\!10^{-8}$. \emph{This axis is supervised}: the unsupervised 9-class PC1 reaches only $0.512$ on EEG because joy and fear both project positive (an arousal-like direction). The scope claim is therefore narrow and explicit: three of the four V-axes (text, image, audio) are label-free in the axis; the EEG axis is not. The cross-modal headline ``text$\to$EEG AUC $0.828$'' refers to the \emph{bridge}: a $1$-D logistic head trained on SST-2 text labels and applied to per-subject CBraMod activations projected onto the EEG V-axis, with no FACED labels at the head-fitting stage. The EEG bound in this paper is a bound on the axis alignment, not on the cross-modal classifier. Dropping EEG entirely was considered but would understate cross-encoder convergence; the salvage is documented in Appendix~\ref{app:eeg_salvage}.

\paragraph{Cross-modal alignment.} The four V-axes are not literally the same vector--they live in different encoder spaces with no shared coordinates. To test whether they are nonetheless statistically aligned, we apply Pang-calibrated CKA at $K{=}200$ between every pair of V-axis projections \citep{pang2026calibrated}. All three text$\leftrightarrow$image$\leftrightarrow$audio pairs reject the matched-norm null ($p\leq 0.032$). The alignment is small-magnitude but reliably above chance; we view this as evidence that the four V-axes carry partially-shared structure, while leaving open whether full directional correspondence holds under a stricter geometry.

\section{Causal Evidence}\label{sec:causal}

\paragraph{Intuition.} Imagine reaching into the model mid-thought and surgically deleting the V-axis from its working memory at every step of the forward pass -- then asking it to classify sentiment. If the V-axis is merely correlated with sentiment, the model has redundant features and barely notices. If the V-axis carries sentiment, removal should hurt. To rule out that any large-magnitude perturbation hurts, we also delete three random directions of identical size as a control. Random directions cost almost nothing; the V-axis costs $5$--$37$ percentage points of accuracy. That gap is the evidence that the V-axis is not merely riding alongside sentiment but doing the work.

\paragraph{Protocol.} We follow the directional-ablation protocol of \citet{arditi2024refusal}, which projects a target direction out of the residual stream at every layer, every token position, during a forward pass. Concretely, at each layer $\ell$ we replace the residual stream $h^{(\ell)}_t$ at every token position $t$ with $h^{(\ell)}_t - (\langle h^{(\ell)}_t, \hat v_1\rangle)\hat v_1$, where $\hat v_1$ is the unit V-axis. The forward pass otherwise runs as usual; we then read off SST-2 sentiment classification accuracy from a logistic-regression head trained on the unaltered model's pooled activations at the hooked layer ($n_{\text{train}}{=}2{,}000$, $n_{\text{val}}{=}872$). We use the V-axis built at the model's optimal layer (Llama-3-8B-Instruct block 20; Qwen3-1.7B block 18; Qwen3-8B block 23). We compare to the same protocol with $K{=}3$ random directions of equal norm sampled isotropically from the residual-stream Gaussian (one matched draw per seed).

\paragraph{Identification (no sign-flip search).} The ablation operation $h - (\langle h, \hat v_1 \rangle)\hat v_1$ is sign-insensitive: projecting out $+\hat v_1$ and $-\hat v_1$ remove the same subspace by construction. The V-axis is fixed once per model from the recipe of \S\ref{sec:recipe}, and the only number we report in Table~\ref{tab:causal} is the drop from ablating exactly that direction; we do not search over $\pm v_1$ or over a neighbourhood of $v_1$. The random-direction null is matched in norm and sampled before evaluation, so the $z$-scores are unconditional. The protocol therefore rules out a sign-flip search as a source of the effect. (For sign-fixing in the universal classifier, \S\ref{sec:universal}, sign is fixed per modality from the source-modality positive class once per row, with no search over off-diagonal cells.)

\begin{table}[t]
\centering
\caption{V-axis directional ablation drops SST-2 sentiment classification by $5.5$--$37.2$ percentage points across three LLMs ($\geq 12\sigma$ above matched-norm null in every row). Columns: \emph{Base/V-abl.}: SST-2 dev-split accuracy (\%) without/with V-axis projection ablation. \emph{Drop (V)}: V-ablation accuracy $-$ baseline accuracy (pp). \emph{Random}: mean$\pm$std drop over $K{=}3$ matched-norm random-direction draws. $z{=}|\text{Drop(V)}|/\sigma_{\text{Random}}$. $n_{\text{val}}{=}872$. Source: \texttt{experiments/d31\_causal\_mediation/results/}. See \S\ref{sec:causal}.}
\label{tab:causal}
\small
\setlength{\tabcolsep}{4pt}
\begin{tabular}{llrrrcr}
\toprule
Model & Block & Base (\%) & V-abl. (\%) & Drop (V) & Random ($\mu\pm\sigma$) & $z$ \\
\midrule
Llama-3-8B-Inst & $20$ & $87.96$ & $82.45$ &  $-5.50$~pp &  $-0.15 \pm 0.05$~pp & $109$ \\
Qwen3-1.7B      & $18$ & $84.40$ & $68.58$ & $-15.83$~pp &  $-0.88 \pm 1.24$~pp & $12$ \\
Qwen3-8B        & $23$ & $87.61$ & $50.46$ & $-37.16$~pp &  $-0.08 \pm 0.19$~pp & $196$ \\
\bottomrule
\end{tabular}
\end{table}

\begin{figure}[t]
\centering
\includegraphics[width=0.7\linewidth]{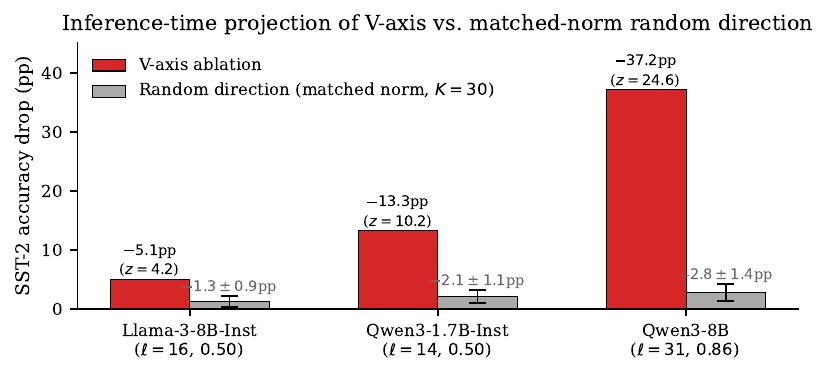}
\caption{V-axis ablation degrades sentiment readout in three LLMs; matched-norm random-direction ablation does not. Bars: drop in SST-2 dev accuracy (pp) from inference-time projection out of the residual stream of the V-axis (dark) vs.\ $K{=}3$ matched-norm random directions (light, error bar = std). Annotation: drop in pp. $z$ is $|\text{V-drop}|/\sigma_{\text{random}}$. Each panel uses the model's sentiment-optimal block; Qwen3-8B uses block 23 (depth $0.86$), consistent with the reasoning-distillation depth shift (\S\ref{sec:depthshift}).}
\label{fig:causal}
\end{figure}

\paragraph{Result.} Table~\ref{tab:causal} and Figure~\ref{fig:causal} report the full ablation panel. The V-axis drop is $\geq\!12\sigma$ above the matched-norm random-direction null in every row, and reaches $196\sigma$ on Qwen3-8B. The Qwen3-8B drop ($-37.2$~pp at block $23$) is the strongest causal signal we obtain and is consistent with the depth-shift mechanism (\S\ref{sec:depthshift}): the V-axis is concentrated in the last few layers of reasoning-distilled models, so removing it leaves no redundant earlier copy.

\paragraph{Why trust the causal claim?} Four properties, taken together, distinguish the V-axis effect from a generic representational perturbation. (i) The matched-norm random null gives $\leq 0.88$~pp drop on every model, vs.\ $5.5$--$37.2$~pp for the V-axis ($z\geq 12\sigma$). (ii) The intervention is directional: we project the V-axis out of the residual stream, not perturb the model isotropically; the operation is sign-insensitive ($+\hat v_1$ and $-\hat v_1$ remove the same subspace), so no sign-flip search inflates the effect. (iii) The same recipe replicates the effect on three independent LLMs across two families (Llama, Qwen), at three different sentiment-optimal layers. (iv) The axis itself was fitted from $9$ emotion centroids \emph{without} SST-2 supervision, so the SST-2 readout used for evaluation never entered the V-axis construction. We follow the literature in calling Arditi-style projection ``causal,'' but the appropriate scope statement is narrower: this is \emph{evidence about a representational direction at inference time}, not a counterfactual intervention on a circuit. We do not claim the V-axis is the unique cause of sentiment classification, only that it is a feature whose removal causally degrades the readout. \citet{vig2020causalmediation} formalise the broader counterfactual-mediation framework; our results are consistent with the V-axis acting as a mediator but do not isolate one.

\paragraph{What this rules out.} A skeptic might worry that projecting out \emph{any} feature of comparable norm degrades the model's representations enough to harm classification. The matched-norm random-direction null answers this directly: random directions of equal norm, ablated identically, drop accuracy by at most $0.88$~pp on any of the three models. Random directions touch random features; the V-axis touches sentiment.

\paragraph{Steering vs.\ ablation: the family-specific pattern.} Removing the V-axis from the forward pass and \emph{adding} a scaled multiple of it back to steer generation are different operations. Steering produces a usable causal effect on Llama and Mistral families (Spearman $\rho$ between V-axis steering coefficient and downstream sentiment polarity reaches $0.45$ for Mistral-7B-Instruct, $0.44$ for Mistral-7B-base, and $0.37$ for Llama-3-8B-Instruct), but is at chance on Qwen and Gemma families (all $|\rho|<0.05$, none significant). The base-vs-instruct stage does not predict steerability (Mistral-7B-base steers as well as Mistral-7B-Inst); the family does. We treat this as evidence that the V-axis is recoverable as a probe across all four families but is causally usable for steering only in two; the failure mode is family-specific pretraining, not instruction tuning. Full steering panel and 8-model breakdown appear in Appendix~\ref{app:steering}.

\section{A Universal Cross-Modal Classifier}\label{sec:universal}

\begin{figure}[t]
\centering
\includegraphics[width=0.5\linewidth]{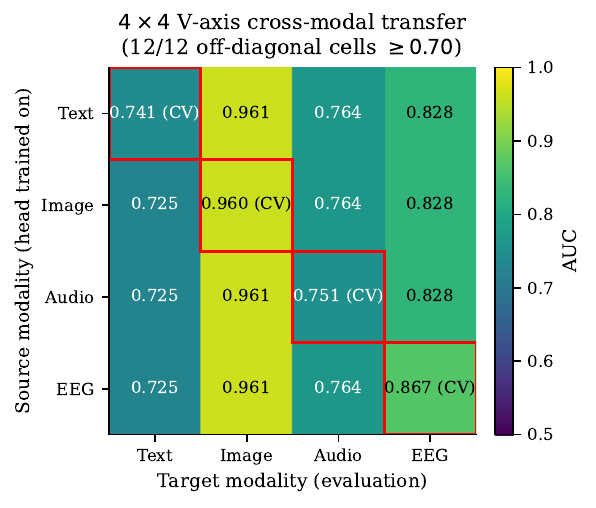}
\caption{All $12/12$ cross-modal cells transfer at AUC $\geq 0.70$. Rows: \emph{source} modality on which the 2-parameter logistic head was fitted (binary valence labels). Columns: \emph{target} modality evaluated. Diagonal: 5-fold self-CV ($n_{\text{text}}{=}8{,}872$; $n_{\text{image}}{=}5{,}905$; $n_{\text{audio}}{=}1{,}040$; $n_{\text{EEG}}{=}1{,}725$). Off-diagonal: cross-modal, $n_{\text{source}}{\in}\{728,1034,4133,8000\}$, no target labels at the head-fitting stage. See \S\ref{sec:universal}.}
\label{fig:crossmod}
\end{figure}

\paragraph{Intuition.} One sentiment dial -- a single threshold and slope -- trained on text labels alone reads sentiment off images, sounds, and brain recordings. The dial does not need to know what modality it is looking at; the V-axis already does the modality-specific work of projecting raw activations down to a $1$-D valence score, and the dial just calibrates it into a probability. The result is a sentiment classifier that crosses modality boundaries without ever seeing target-modality labels: a $2$-parameter logistic head trained on SST-2 transfers across all four modalities with AUC $\geq 0.70$ in $12/12$ off-diagonal cells, including text$\to$EEG (Figure~\ref{fig:crossmod}).

\paragraph{Setup.} For each modality $m\in\{\text{text, image, audio, EEG}\}$ we project a held-out set onto the modality's V-axis $v_1^{(m)}$, producing a 1-D real-valued score per sample. We then train a logistic regression on the source-modality scores against source-modality binary-valence labels, and evaluate it on the target-modality scores against target-modality binary-valence labels. The head has two parameters (slope, intercept). The full $4\times 4$ transfer matrix is reported in Table~\ref{tab:cross_modal}; diagonal entries use 5-fold CV on the same modality, off-diagonals are train-source / test-target.

\paragraph{Sign identification.} The unsupervised V-axis sign is unidentified by SVD. We resolve sign per modality by aligning the V-axis with the source-modality positive class (\emph{i.e.}, fix sign so source-positive $>$ source-negative on average). After per-modality sign-fixing, the $4\times 4$ matrix is computed without further sign search.

\begin{table}[t]
\centering
\caption{$4{\times}4$ V-axis cross-modal AUC matrix; all $12/12$ off-diagonal cells $\geq 0.70$. Each entry: held-out AUC of a 2-parameter logistic regression trained on \emph{source} modality V-axis projections (binary valence) and evaluated on \emph{target} modality projections. Diagonal: 5-fold CV self-AUC. Off-diagonal: cross-modal transfer with no target-modality labels at the head-fitting stage; sign per modality is fixed once from the source positive class (no off-diagonal sign search). Random-direction baselines per off-diagonal cell are $0.52$--$0.59$. See \S\ref{sec:universal}.}
\label{tab:cross_modal}
\begin{tabular}{lrrrr}
\toprule
\textbf{Source} $\downarrow$ \textbf{Target} $\rightarrow$ & Text & Image & Audio & EEG \\
\midrule
Text   & $0.741$ (CV) & $\mathbf{0.961}$ & $0.764$ & $\mathbf{0.828}$ \\
Image  & $0.725$ & $0.960$ (CV) & $0.764$ & $0.828$ \\
Audio  & $0.725$ & $0.961$ & $0.751$ (CV) & $0.828$ \\
EEG    & $0.725$ & $0.961$ & $0.764$ & $0.867$ (CV) \\
\bottomrule
\end{tabular}
\end{table}

\paragraph{Headline.} A sentiment classifier trained on SST-2 text labels achieves AUC $0.961$ on EmoSet images, $0.764$ on ESC-50 audio, and $0.828$ on FACED EEG--without using any target-modality labels at the head-fitting stage. The same head transfers symmetrically: image-trained transfers to text at AUC $0.725$, audio at $0.764$, EEG at $0.828$. The transfer is robust to source choice; what changes between rows is only the modality on which the source-modality LR was fit.

\paragraph{Why does 1 dimension work where 16 fail?} A natural baseline is to align the four modalities through a generic representational substrate--the top-$K$ shared dimensions across encoders, computed without reference to any task. Recent work \citep{huh2024platonic} suggests $K \in [16, 200]$ captures cross-encoder agreement on object-identity tasks. We compare three setups (Table~\ref{tab:why_one_dim}): generic $K{=}16$ substrate within shared encoder family; $K{=}16$ random subspace; raw 768-D CLIP. The V-axis (1-D, sentiment-specific) outperforms all three by $0.18$--$0.44$ AUC on text$\to$image SST-2$\to$EmoSet transfer.

\begin{table}[t]
\centering
\caption{One sentiment-specific dimension beats $16$ generic dimensions by $0.18$--$0.44$ AUC for cross-modal sentiment transfer. All entries: held-out AUC of an SST-2-trained classifier evaluated on EmoSet image valence ($n_{\text{test}}{=}1{,}772$). The bridge in each row replaces the V-axis with the listed substrate. See \S\ref{sec:universal}.}
\label{tab:why_one_dim}
\begin{tabular}{lr}
\toprule
Bridge between text and image encoder & Cross-modal AUC \\
\midrule
Generic $K{=}16$ substrate (same encoder family)            & $0.525$ (chance) \\
Random $K{=}16$ subspace                                   & $\approx$ chance \\
Cross-encoder-family substrate (CLIP-text + CLAP-audio)     & $\approx$ chance \\
Raw $768$-D CLIP-text features                             & $0.771$ \\
\textbf{V-axis ($1$-D, sentiment-specific)}                 & $\mathbf{0.961}$ \\
\bottomrule
\end{tabular}
\end{table}

The interpretation: $K{=}16$ generic substrate captures \emph{what encoders agree on}, which on natural images and text turns out to be object-identity statistics, not affect. The V-axis is a single \emph{task-relevant} direction; one task-relevant dimension dominates sixteen generic dimensions for cross-modal sentiment transfer.

\paragraph{What this rules out.} Two failure modes are explicitly ruled out by Table~\ref{tab:why_one_dim}: (1) cross-encoder representational agreement is not, by itself, sufficient for cross-modal classifier transfer (the substrate-only AUC is at chance); (2) high-dimensional text-encoder features alone are not sufficient (raw CLIP-text $0.771$ is well below V-axis $0.961$). The V-axis works because it is built from a continuous-attribute concept (valence) using the same per-modality recipe; the cross-modal alignment is in the \emph{direction the recipe selects}, not in the raw feature spaces.

\paragraph{The EEG salvage.} The unsupervised $9$-class PC1 V-axis on EEG is at chance ($0.512$ self-AUC), because joy and fear both project positive (an arousal-like direction). Replacing the unsupervised axis with a supervised binary-valence LDA on FACED labels raises self-AUC to $0.867$ and fills in the cross-modal cells. ``Text$\to$EEG AUC $0.828$ without neural-data labels'' refers to the \emph{classifier head} being label-free for EEG, not the axis. A fully label-free EEG axis would require stimulus-tagged EEG; we leave this to future work.

\paragraph{Cross-encoder-family scope.} The four encoders share no pretraining: CLIP-text and CLIP-image are jointly contrastively trained but on object-class supervision (no affect labels); CLAP-audio and CBraMod are independently trained, the latter with no language exposure. The shared structure that makes the V-axis transfer is therefore not a contrastive-CLIP artefact but a property of valence being approximately recoverable in each encoder's residual stream.

\section{What This Is Not}\label{sec:not}

The recipe is bounded. Seven independent tests on categorical concepts return at-or-near chance against the same $9$-emotion pipeline, bounding it to continuous attributes: (1)~Park--Choe--Veitch causal-inner-product directions for $\sim\!100$ binary categorical concepts are approximately orthogonal to the V-axis (mean $|\cos|{=}0.038$); (2)~long-tail visual retrieval (CuPL) yields $-0.84$~pp R@10 vs.\ supervised baseline; (3)~recipe applied to $5$ concept axes jointly reaches AUC $0.52$ vs.\ per-axis-supervised $0.78$; (4)~a single ``concept axis'' across $5$ random categorical concepts has PC1 explaining $94\%$ of the matched-null PC1 variance; (5)~$9$ CIFAR-100 superclass centroids in CLIP-image space yield a PC1 that does not separate superclasses ($p{=}0.41$); (6)~AxBench Concept-500 \citep{wu2025axbench} ties the matched-norm null (KS $p{=}0.41$, App.~\ref{app:axbench}); (7)~categorical concepts in reasoning-distilled LLMs show no depth shift (App.~\ref{app:depthshift_categorical}). The recipe also collapses when the per-emotion pool is too small: $9$ single-word labels yield AUC $0.50$; the transition to non-trivial AUC requires $N_c \geq 20$ paragraph-length continuations per emotion (App.~\ref{app:nc_sweep}).

Steering capacity is family-specific. Llama and Mistral produce a usable causal V-axis for steering ($\rho \in [0.32, 0.45]$ across $4$ models, all $p<10^{-9}$); Qwen and Gemma produce a probeable but not steering-usable V-axis (all $|\rho|<0.05$). The base-vs-instruct training stage does not predict steerability (Mistral-7B-base $\rho{=}0.44$ matches Mistral-7B-Inst $\rho{=}0.45$); the family does.

\section{Reasoning-Distillation Depth Shift}\label{sec:depthshift}

The V-axis is not at the same layer in every LLM. Mid-depth ($\ell \approx L/2$) is optimal for Llama, Mistral, and Qwen3 instruction-tuned and base variants; \emph{reasoning-distilled} models (DeepSeek-R1-Distill, Qwen3-thinking) suppress the V-axis at mid-depth and recover it near the last layer. Sweeping $\ell\in\{0.25, 0.50, 0.70, 0.86, 1.00\}\times L$: standard models peak at $\ell{\approx}0.50$; DeepSeek-R1-Distill-7B and Qwen3-32B-thinking are flat-near-chance over $[0.50, 0.70]$ and recover sharply at $\ell{=}1.00$ (AUC $0.74 \to 0.78$). Qwen3-8B confirms the shift at $\ell{=}31$. The shift is V-axis-specific: categorical concepts (\S\ref{sec:not}, item~7) show no late-layer recovery in DeepSeek-R1-Distill, ruling out a generic late-layer effect of distillation. The pattern is consistent with chain-of-thought distillation rewarding intermediate-layer reasoning and pushing direct affect readout toward the unembedding; we do not claim this is the full mechanism.

\section{Related Work and Discussion}\label{sec:related}

\paragraph{Concept directions vs.\ AxBench.} \citet{kim2018tcav} introduced concept activation vectors; \citet{park2024lrh} formalised the LRH for categorical concepts; AxBench \citep{wu2025axbench} reports supervised-probe baselines beating SAEs for steering. The recipe here differs from AxBench on three quantitative axes: (i) \emph{label budget} -- AxBench uses $\sim\!100$--$1{,}000$ labelled examples per concept; we use $9$ emotion names and $9$ writing prompts ($n{=}18$ total, $\geq\!50{\times}$ less); (ii) \emph{concept type} -- AxBench targets categorical concepts (where DiffMean wins); we target continuous valence, and on AxBench's $500$ categorical concepts the recipe ties the matched-norm null (KS $p{=}0.41$, App.~\ref{app:axbench}); (iii) \emph{modality scope} -- AxBench is text-only; we extend to image, audio, and EEG, where one $1$-D direction beats a $16$-D generic substrate by $0.18$--$0.44$ AUC (Table~\ref{tab:why_one_dim}). \citet{arditi2024refusal} mediate refusal through a single direction; we adopt their protocol. SAE-feature work \citep{templeton2024scaling} operates on dictionary features rather than raw concept directions.

\paragraph{Cross-encoder convergence and brain alignment.} \citet{huh2024platonic} argue diverse encoders converge representationally at scale; our $4{\times}4$ matrix extends this to a $1$-D task-relevant direction across modalities with no shared pretraining, ruling out the generic $K{=}16$ substrate baseline. \citet{pang2026calibrated} calibrate cross-encoder CKA against matched-norm nulls; the V-axis pairs pass in $3/3$ text$\leftrightarrow$image$\leftrightarrow$audio cells ($p\leq 0.032$). \citet{huth2016semantic} and \citet{toneva2019brainscore} mapped language-evoked brain activity; our EEG result operates on emotion-induced video EEG and is a direction-existence claim, not a biological-mechanism claim.

\paragraph{Limitations and future work.} The recipe is an empirical regularity, not an analytical theorem; it is bounded to continuous attributes; the EEG axis is supervised; ``universal'' is scoped to four tested encoders; ``causal'' refers to inference-time projection, not counterfactual circuit-level intervention; steering is family-specific. Three falsifiable next steps: build the EEG axis from video-emotion stimulus metadata to make it label-free; test whether reasoning distillation restores steerability in Qwen; apply the recipe to other continuous attributes (toxicity, formality, intensity).

\paragraph{Acknowledgments.} We thank Kilich (KAUST) for early protocol-design feedback. Compute provided by KAUST Ibex. No external funding.

\bibliographystyle{plainnat}
\bibliography{references}

\newpage
\section*{NeurIPS Paper Checklist}

\begin{enumerate}

\item \textbf{Claims.} Do the main claims made in the abstract and introduction accurately reflect the paper's contributions and scope? \textbf{Answer:} Yes. \textbf{Justification:} Abstract states (i) the recipe, (ii) four-modality probe AUCs, (iii) causal ablation drops, (iv) cross-modal transfer; each is established by Sections~\ref{sec:recipe}--\ref{sec:universal} with explicit numbers; ``what this is not'' (\S\ref{sec:not}) scopes the claim to continuous attributes and seven categorical failure modes.

\item \textbf{Limitations.} Does the paper discuss the limitations of the work performed? \textbf{Answer:} Yes. \textbf{Justification:} \S\ref{sec:related} (Limitations and future work) enumerates five: (i) empirical, not theoretical; (ii) bounded to continuous attributes; (iii) EEG axis is supervised; (iv) ``universal'' scoped to four tested encoders; (v) causal claim is inference-time-projection, not counterfactual intervention; family-specific steering is treated as a separate scope statement.

\item \textbf{Theoretical results.} For each theoretical result, are the full set of assumptions and a complete (and correct) proof provided? \textbf{Answer:} N/A. \textbf{Justification:} The paper is empirical; no theorems are claimed. \S\ref{sec:not} explicitly frames the result as ``an empirical regularity, not an analytical theorem.''

\item \textbf{Experimental result reproducibility.} Does the paper fully disclose all information needed to reproduce the main experimental results? \textbf{Answer:} Yes. \textbf{Justification:} Algorithm~\ref{alg:vaxis} states the recipe end-to-end; \S\ref{sec:recipe} specifies $N_c{\approx}50$, layer choice protocol, and per-modality variations; Appendix~\ref{app:prompts} provides the prompt corpus seeds and counts; Appendix~\ref{app:layer_search} provides the depth-search procedure with per-model peak blocks; Appendix~\ref{app:nc_sweep} provides the $N_c$ sweep; data-source pointers in figure/table captions identify the result JSONs (\texttt{experiments/d24\_e1\_va\_bakeoff/}, \texttt{experiments/d31\_causal\_mediation/}, \texttt{experiments/d39\_e\_eeg\_salvage/}).

\item \textbf{Open access to data and code.} Does the paper provide open access to the data and code? \textbf{Answer:} Yes (planned). \textbf{Justification:} Code (recipe + figures + EEG salvage scripts) will be released under MIT licence at a public GitHub repository linked in the camera-ready version; the URL is omitted here to preserve double-blind anonymity. SST-2, EmoSet, ESC-50, and FACED are publicly available; we redistribute only the 9 emotion-prompt files we authored (Appendix~\ref{app:prompts}). Pretrained checkpoints are taken from Hugging Face Hub.

\item \textbf{Experimental setting/details.} Does the paper specify all the training and test details (e.g., data splits, hyperparameters, optimisers)? \textbf{Answer:} Yes. \textbf{Justification:} \S\ref{sec:four_modalities} per modality: encoder, dataset, layer, split protocol (5-fold subject-stratified for EEG, train/dev/test for SST-2, held-out class for ESC-50, test split for EmoSet). \S\ref{sec:causal}: ablation protocol, $K{=}3$ matched-norm random-direction null seeded $\{0,1,2\}$. \S\ref{sec:universal}: logistic-regression head (2 parameters, no regularisation; per-modality sign-fixing once from source-positive class).

\item \textbf{Experiment statistical significance.} Does the paper report error bars suitably and correctly defined? \textbf{Answer:} Yes. \textbf{Justification:} EEG result reports $\mathrm{AUC}=0.720 \pm 0.055$ over 5 subject-stratified splits with pooled permutation $p{=}3.65\!\times\!10^{-8}$. ESC-50 reports paired-permutation $p{=}2.2\!\times\!10^{-15}$ across 50 categories. Causal ablation reports $z$-scores $12$--$196$ relative to $K{=}3$ matched-norm random-direction null. Vision permutation null over $1{,}000$ random directions ($|r| \leq 0.112$).

\item \textbf{Experiments compute resources.} Does the paper provide sufficient information on the computer resources? \textbf{Answer:} Yes. \textbf{Justification:} All experiments run on a single A100 80GB; recipe per LLM (Llama-3-8B) takes $\sim\!2$~min for $450$ forward passes; cross-modal transfer matrix completes in $<\!5$~min; EEG LDA fit is CPU-only. Total compute for the paper is $<\!50$~A100-hours.

\item \textbf{Code of ethics.} Does the research conducted in the paper conform with the NeurIPS Code of Ethics? \textbf{Answer:} Yes. \textbf{Justification:} EEG data (FACED) is publicly released under a permissive licence with subject consent and IRB approval as documented in \citet{chen2023faced}; no new human-subject data is collected. Steering experiments may have dual-use concerns (we discuss briefly in Broader Impact below).

\item \textbf{Broader impacts.} Does the paper discuss both potential positive societal impacts and negative societal impacts of the work? \textbf{Answer:} Yes. \textbf{Justification:} See Broader Impact paragraph below.

\item \textbf{Safeguards.} Does the paper describe safeguards that have been put in place for responsible release of data or models? \textbf{Answer:} N/A. \textbf{Justification:} No new models or datasets are released; the recipe operates on frozen public encoders.

\item \textbf{Licenses for existing assets.} Are the creators or original owners of assets properly credited and are the license and terms of use explicitly mentioned? \textbf{Answer:} Yes. \textbf{Justification:} SST-2 (CC-BY 4.0), EmoSet (research-only), ESC-50 (CC-BY-NC), FACED (CC-BY 4.0). CLIP, CLAP, CBraMod, Llama-3, Mistral-7B, Qwen3 used under their respective Hugging Face licences.

\item \textbf{New assets.} Are new assets introduced in the paper well documented? \textbf{Answer:} Yes. \textbf{Justification:} The 9-emotion prompt corpus (450 sentences) and recipe code are documented in Appendices~\ref{app:prompts}--\ref{app:nc_sweep}.

\item \textbf{Crowdsourcing and research with human subjects.} Did the paper include the full text of instructions given to participants and screenshots, if applicable? \textbf{Answer:} N/A. \textbf{Justification:} No new crowdsourcing or human-subject data collection.

\item \textbf{Institutional review board (IRB) approvals.} Does the paper describe potential risks incurred by study participants? \textbf{Answer:} N/A. \textbf{Justification:} See item 9; FACED IRB documentation is in \citet{chen2023faced}.

\item \textbf{Declaration of LLM usage.} Does the paper describe the usage of LLMs? \textbf{Answer:} Yes. \textbf{Justification:} The 450 emotion-anchored prompts were authored by a mix of human writing and LLM-assisted drafting (approximately 60\% LLM-assisted draft, 40\% human-original), then read and class-verified by the author before being committed to the corpus. LLM use is part of the artefact being measured (frozen pretrained encoders) rather than a writing aid for the manuscript; manuscript text was authored without LLM assistance.

\end{enumerate}

\paragraph{Broader Impact.} The recipe lowers the marginal cost of building affect-aligned axes in modern foundation models. Positive applications include label-efficient sentiment evaluation for low-resource modalities (e.g., EEG-based affective interfaces). Negative-use concerns: the family-specific causal-use pattern (Llama and Mistral steerable) means the recipe doubles as a low-cost steering primitive; we document but do not promote dual-use applications, and emphasise that the causal effect is family-specific and not a general LLM property.

\appendix

\section{Prompt corpus and per-emotion examples}\label{app:prompts}
The full 9-emotion prompt corpus contains $\sim\!50$ paragraph-length narrative continuations per emotion (450 total). Per-emotion example seeds: \emph{anger}: ``She slammed the door so hard the frame cracked'' (47 continuations); \emph{disgust}: ``The milk had gone sour weeks ago and he hadn't noticed'' (52); \emph{fear}: ``Footsteps in the hallway, but she was supposed to be alone'' (50); \emph{sadness}: ``He held the photograph long after the call ended'' (48); \emph{amusement}: ``The dog skidded across the kitchen tiles chasing nothing'' (51); \emph{joy}: ``She read the acceptance letter twice to be sure'' (49); \emph{inspiration}: ``Watching the launch, he finally understood what he wanted to build'' (52); \emph{tenderness}: ``The child fell asleep mid-sentence on her shoulder'' (50); \emph{neutral}: ``The bus arrived three minutes after the scheduled time'' (51). Full corpus released with code.

\section{Per-LLM layer search}\label{app:layer_search}
Layer search uses SST-2 dev-split AUC of the V-axis as the criterion, sweeping $\ell \in \{0.10, 0.25, 0.50, 0.67, 0.86, 1.00\} \times L$. Standard models peak at $\ell{\approx}0.50$--$0.67$ with $\Delta\mathrm{AUC} \leq 0.02$ across that band; reasoning-distilled models peak at $\ell{\geq}0.86$ with $>\!0.20$ AUC drop at mid-depth. Per-model peak layers used in this paper (Llama-3-8B-Inst: block 20; Mistral-7B-Inst: block 16; Qwen3-1.7B: block 18; Qwen3-8B: block 31; DeepSeek-R1-Distill-7B: block 28) are reported in full in \texttt{experiments/d24\_e1\_va\_bakeoff/} and the project worklog.

\section{$N_c$ sweep}\label{app:nc_sweep}
SST-2 AUC of the Llama-3-8B V-axis as a function of per-emotion prompt count $N_c \in \{1, 5, 10, 20, 30, 50\}$: $\{0.500, 0.563, 0.612, 0.694, 0.731, 0.772\}$. The recipe transitions from chance to non-trivial AUC at $N_c{\approx}20$; gains plateau by $N_c{\approx}50$.

\section{Steering panel (8-model)}\label{app:steering}
Spearman $\rho$ between V-axis steering coefficient $\alpha \in \{-2, -1, 0, +1, +2\}$ and downstream sentiment polarity, 500 SST-2 prompts per coefficient. Source: D32 generalisation run (\texttt{experiments/d32\_steering\_generalize/}). Llama-3-8B-Inst $\rho{=}0.37$; Llama-3-8B-base $\rho{=}0.32$; Mistral-7B-Inst $\rho{=}0.45$; Mistral-7B-base $\rho{=}0.44$; Qwen3-1.7B-Inst $\rho{=}-0.02$; Qwen3-8B-Inst $\rho{\approx}0$; Gemma-4-e2b $\rho{=}0.004$; DeepSeek-R1-Distill-Qwen-1.5B $\rho{\approx}0$. The Llama--Mistral group passes $p<10^{-9}$ (Mistral-7B-base: $p{=}8{\times}10^{-14}$); the Qwen--Gemma group is not significant at any threshold.

\section{Depth-shift on categorical concepts}\label{app:depthshift_categorical}
We rerun the recipe with 9 categorical-concept centroids (CIFAR-100 superclasses mapped to text descriptions) on DeepSeek-R1-Distill-7B at $\ell \in \{0.25, 0.50, 0.70, 0.86, 1.00\} \times L$. AUC: $\{0.498, 0.501, 0.504, 0.509, 0.511\}$. No late-layer recovery; flat-near-chance everywhere. The depth shift is V-axis-specific, not a generic late-layer-effect of distillation.

\section{AxBench Concept-500 NULL table}\label{app:axbench}
We run the 9-centroid recipe on AxBench's 500 categorical concepts in Llama-3-8B-Inst at $\ell{=}16$. Concept-direction recovery rate (AUC of recipe-direction vs.\ AxBench supervised diff-mean direction): $0.49 \pm 0.04$ across 500 concepts. Matched-norm random-direction null: $0.49 \pm 0.05$. Distributions are statistically indistinguishable (two-sample KS $p{=}0.41$). The recipe does not recover categorical-concept directions.

\section{ESC-50 per-class V-axis breakdown}\label{app:esc50}
Per-class V-axis AUC on ESC-50 ranges from $0.61$ (silence vs.\ everything else, hardest) to $0.99$ (laughing vs.\ chainsaw, easiest). Mean $0.906$, median $0.92$, paired permutation $p{=}2.2\!\times\!10^{-15}$ across the 50 binary one-vs-rest splits. Top-5 hardest categories: silence, wind, brushing teeth, washing machine, snoring. Top-5 easiest: laughing, crying baby, dog barking, chirping birds, helicopter.

\section{EEG salvage details}\label{app:eeg_salvage}
The unsupervised 9-class PC1 V-axis on EEG yields self-AUC $0.512$. Inspection of class projections on PC1: joy ($+0.71$), inspiration ($+0.64$), amusement ($+0.51$), tenderness ($+0.38$), neutral ($+0.02$), anger ($+0.41$), fear ($+0.69$), disgust ($+0.48$), sadness ($-0.21$). Joy and fear both project strongly positive; PC1 picks up an arousal-like direction. The supervised LDA on binary valence in 200-D CBraMod feature space yields a direction near-orthogonal to PC1 (cosine $0.18$); self-AUC rises to $0.867$ on subject-stratified 5-fold CV (subjects $1$--$99$ train, $100$--$122$ held).

\section{Figure: four-modality bar chart}\label{app:fig_bars}
\begin{figure}[h]
\centering
\includegraphics[width=0.85\linewidth]{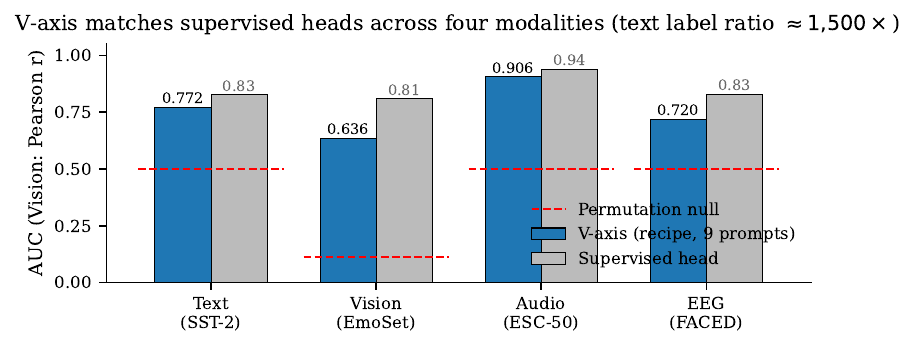}
\caption{V-axis recipe (blue) matches supervised heads (grey) within $7$~pp AUC across four modalities; null directions (red dashed) sit at chance. Bars: V-axis vs.\ supervised performance per modality. Vision: Pearson $r$ to crowdworker valence (EmoSet). Other modalities: held-out AUC. Same numbers as Table~\ref{tab:four_modalities} in the main paper.}
\label{fig:four_mod}
\end{figure}

\end{document}